\documentclass{bmvc2k}
\usepackage{graphicx}
\usepackage{tikz}
\usepackage{comment}
\usepackage{amsmath,amssymb} 
\usepackage{color}

\usepackage{bm}
\usepackage{multirow}
\usepackage{xcolor}

\usepackage{booktabs}

\title{Inharmonious Region Localization with Auxiliary Style Feature}

\addauthor{Penghao Wu}{wupenghaocraig@sjtu.edu.cn}{1}
\addauthor{Li Niu$^\ast$}{ustcnewly@sjtu.edu.cn}{1}
\addauthor{Liqing Zhang}{zhang-lq@cs.sjtu.edu.cn}{1}

\addinstitution{
 MoE Key Lab of Artificial Intelligence \\
 Shanghai Jiao Tong University \\
 Shanghai, China
}

\runninghead{Wu ET AL.}{Inharmonious Region Localization with ASF}

\begin{document}

\maketitle

\begin{abstract}
With the prevalence of image editing techniques, users can create fantastic synthetic images, but the image quality may be compromised  by  the  color/illumination  discrepancy  between  the  manipulated region and background. Inharmonious region localization aims to localize the inharmonious region in a synthetic image. In this work, we attempt to leverage auxiliary style feature to facilitate this task. Specifically, we propose a novel color mapping module and a style feature loss to extract discriminative style features containing task-relevant color/illumination information.  Based  on  the  extracted  style  features,  we  also  propose  a novel style voting module to guide the localization of inharmonious region. Moreover, we introduce semantic information into the style voting module to achieve further improvement. Our method surpasses the existing methods by a large margin on the benchmark dataset.
\end{abstract}

\section{Introduction}
\label{sec:intro}

\begin{figure}[t]
\centering
\includegraphics[scale=0.45]{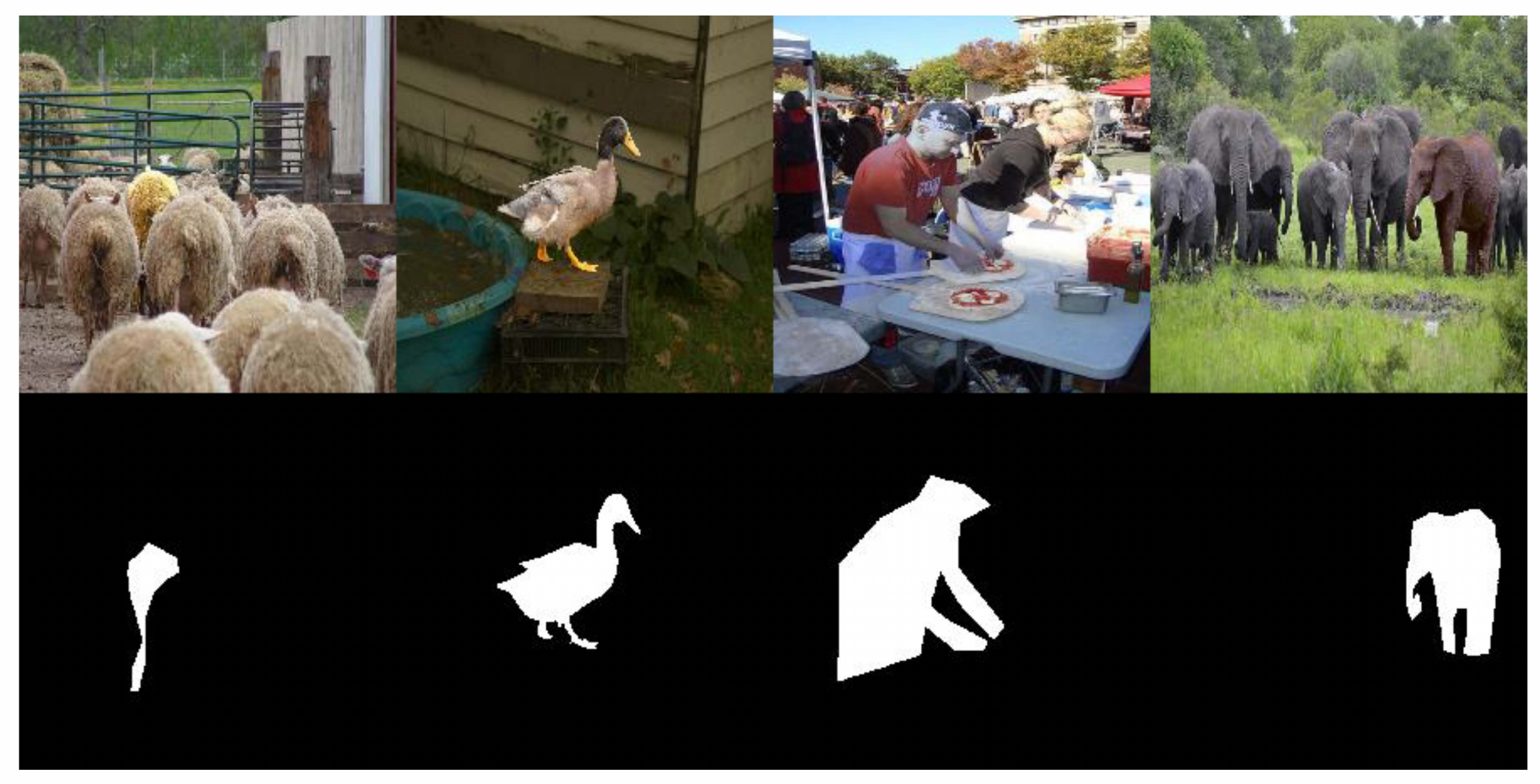}
\caption{Examples of inharmonious images ($1^{st}$ row) and their associated masks ($2^{nd}$ row).}
\label{fig:inharmonious_example}
\end{figure}

With the wide application of photography and editing technology, people can easily create marvellous synthetic images with common image editing  operations (\emph{e.g.}, copy-paste, appearance adjustment). However, one serious problem of synthetic images is that the manipulated region may have inconsistent color and illumination characteristics with the background (see Fig.~\ref{fig:inharmonious_example}), making the whole image inharmonious and unrealistic. The inharmonious region localization task \cite{DIRL} aims to localize the inharmonious region, after which 
users can manually adjust the inharmonious region or utilize image harmonization techniques \cite{tsai2017deep,cong2020dovenet} to harmonize the inharmonious region, yielding the images with higher quality and fidelity.
Therefore, inharmonious region localization is indispensable for blind image harmonization \cite{cun2020improving}, in which the inharmonious region mask is unavailable. 

The first method on inharmonious region localization is DIRL \cite{DIRL}, which mainly focuses on the backbone design to fuse multi-scale features and suppress redundant information, ignoring critical information related to color and illumination statistics, which is actually the essence of this task. Recent work MadisNet~\cite{madisnet} transforms the image to another color-space to magnify domain discrepancy. Different from previous works, we aim to extract \textit{style features} containing task-relevant color and illumination information with a style encoder, to help localize the inharmonious region. By dividing the image into inharmonious region and harmonious (background) region, we enforce intra-region coherence and inter-region divergence using a \textit{style feature loss}, by pulling close the pixel-level style features within the same region while pushing apart those across different regions.
Unlike the complex color mapping model in MadisNet~\cite{madisnet}, we design a simple yet effective \textit{color mapping module} to manipulate the input image to help extract more discriminative style features.

To fully exploit the potential of the pixel-level style features extracted from the style encoder, we propose a novel \textit{style voting module} and insert it into each decoder stage. Each decoder stage produces an auxiliary inharmonious region mask which is used to select harmonious pixels as voters. These voters need to vote for the pixels with similar style features. For each pixel, the total score it receives represents its probability of being harmonious. 
Furthermore, considering that semantically similar regions usually provide more reliable clues, we include extra semantic information to calculate weights, which are assigned to the scores given from each voter to each pixel. Finally, the weighted total scores of all pixels form the voting score map, which serves as the guidance for the next decoder stage. 

In summary, our whole network consists of a style encoder and a UNet structure, which are linked together by style features. A color mapping module is placed in front of style encoder to help extract better style features. A style voting module leverages the style features to produce a style voting map to guide the UNet decoder. Because the semantic information in style voting module is expensive and optional, we refer to our base model without semantic information as AustNet (\textbf{Au}xiliary \textbf{St}yle feature) and the enhanced version with semantic information as AustNet-S.
Following \cite{DIRL}, we conduct experiments on the benchmark dataset iHarmony4 \cite{cong2020dovenet} and our method outperforms the state-of-the-art method MadisNet~\cite{madisnet} by a large margin. Specifically, we improve the AP from 85.86\% to 92.01\% with our AustNet and to 93.01\% with our AustNet-S compared with the SOTA results. Our main contributions include:

    $\bullet$ We design a style encoder with a simple color mapping module to extract discriminative style feature to facilitate inharmonious region localization. 
    
    $\bullet$ We propose a novel style voting module based on extracted style features to provide guiding information for the decoder.
    
    $\bullet$ We introduce semantic information into the weighting scheme in 
    our style voting module, which can achieve further improvement. 

\section{Related Work}
\label{sec:relatedwork}

\subsection{Image Manipulation Detection}
Image manipulation detection aims to detect and localize the manipulated or tampered image, which is somewhat similar to inharmonious region localization. In image manipulation detection,  manipulation operations usually contain copy-move, removal, inpainting, and splicing. 
Traditional image manipulation methods heavily rely on prior information of manipulated images like noise patterns \cite{mahdian2009using,Pun2016MultiscaleNE} and JPEG compression artifacts \cite{5946978,6151134,LI20091821}. Recently, deep learning approaches tackle the image manipulation detection task by comparing local patches \cite{8237794,7900012,7823911}, extracting forgery features \cite{8953774,9102825,zhou2018learning,8335799}, and  adversarial learning \cite{Kikuchi_2019_ICCV}.
However, the discrepancy between color or illumination statistics, which is the main focus of inharmonious region localization, is not specifically studied in these methods.


\subsection{Image Harmonization}
Image harmonization \cite{tsai2017deep,cong2020dovenet,Guo_2021_CVPR,cong2022high,hang2022scs,bao2022deep,CharmNet} aims to adjust the foreground of a composite image in terms of color and illumination characteristics, to make the foreground compatible with the background, which can be deemed as a successor task of inharmonious region localization. Tsai \emph{et al.} \cite{tsai2017deep} proposed the first convolutional neural network for image harmonization. A spatially separated attention module S$^2$AM was proposed in \cite{cun2020improving} to process the features of foreground and background differently. Domain translation was adopted in \cite{cong2020dovenet,9428394} to translate the inharmonious foreground to the background domain. Images are decomposed into illumination map and reflectance map in \cite{Guo_2021_CVPR,Guo_2021_ICCV} to tackle the harmonization problem. Semantic information was utilized in    \cite{sofiiuk2021foreground} to assist with the harmonization process.

Although image harmonization methods have shown impressive performance, most of them require the ground-truth foreground mask, which are not always available in real-world applications. For blind image harmonization without provided foreground mask, S$^2$AM predicts the inharmonious foreground mask as an auxiliary task, but it is not the main focus and the quality is very low. Therefore, the importance of inharmonious region localization is significant and the quality of the predicted mask is crucial for blind image harmonization. 


\subsection{Inharmonious Region Localization}
Inharmonious region localization aims to localize the inharmonious region which is incompatible with the background due to distinctive color and illumination statistics. DIRL \cite{DIRL} is the first method working on inharmonious region localization, which develops an effective way to merge multi-scale features and use mask guided attention module to localize the inharmonious region. However, the discrepancy about color and illumination information was not fully exploited in \cite{DIRL}. MadisNet \cite{madisnet} uses HDRNet \cite{hdrnet} to map the input color space to magnify the domain discrepancy, while we propose to use a simple linear color mapping module to better extract style features. Moreover, we leverage the style features in two aspects, which has never been explored before. We are also the first to introduce semantic information into inharmonious region localization task. 
\section{Methodology}
\label{sec:method}

\begin{figure}[t]
\centering
\includegraphics[scale=0.5]{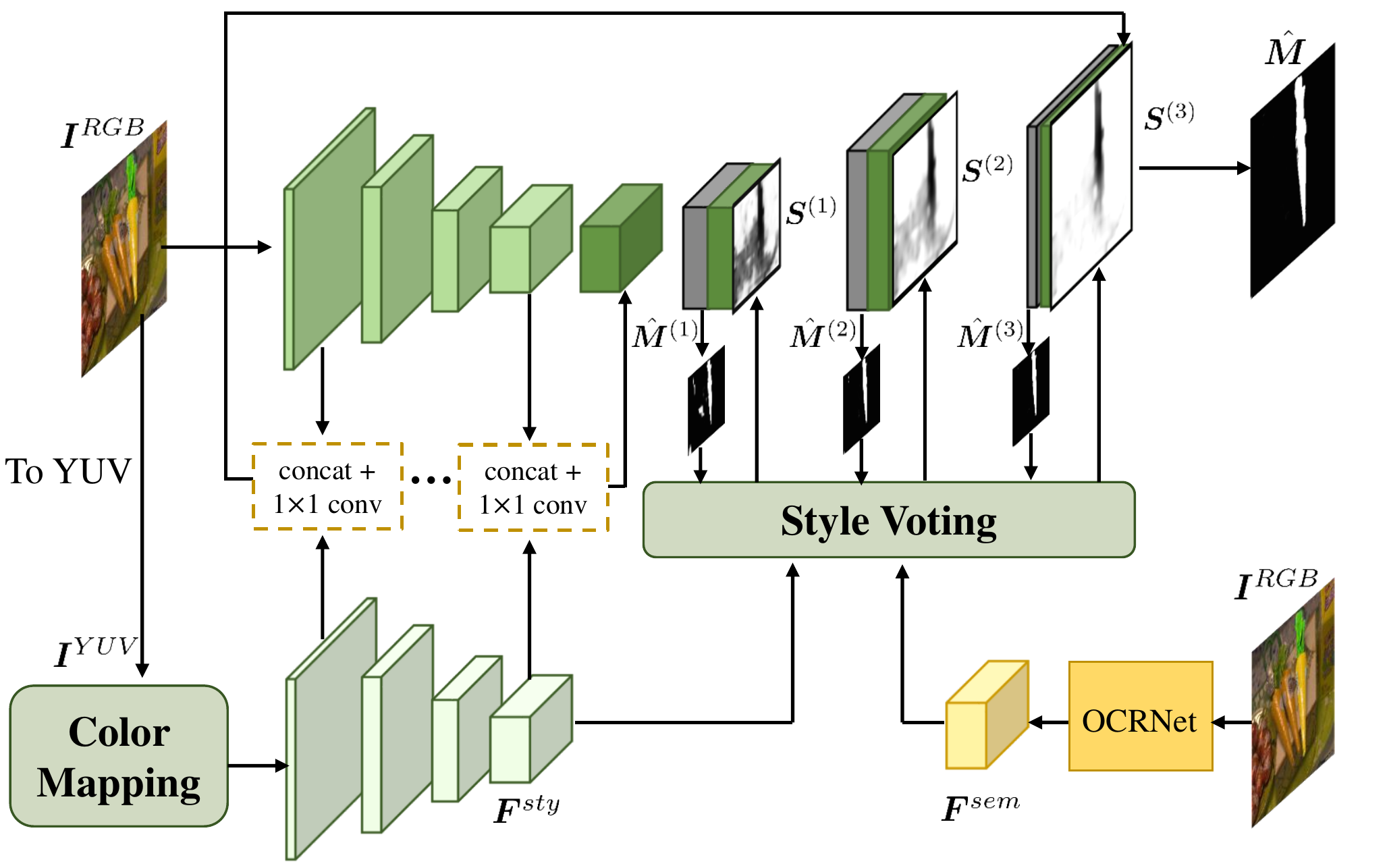}
\caption{Our network structure is comprised of a style encoder and a UNet~\cite{Unet}  structure. We insert a color mapping module in front of style encoder to help extract better style features $\bm{F}^{sty}$. The style voting module takes in $\bm{F}^{sty}$, the semantic features $\bm{F}^{sem}$ from pretrained segmentation network (\emph{e.g.}, OCRNet~\cite{OCRNet}), and the auxiliary inharmonious region mask $\hat{\bm{M}}^{(k)}$ from each decoder stage, producing a style voting map $\bm{S}^{(k)}$.  }
\label{fig:pipeline}
\end{figure}

Given a synthetic RGB image $\bm{I}^{RGB}$, our goal is estimating a binary mask $\hat{\bm{M}}$ to localize the inharmonious region. As shown in Fig.~\ref{fig:pipeline}, our whole framework is comprised of a UNet \cite{Unet} and a style encoder. For UNet, we adopt ResNet34 \cite{ResNet} as the encoder, which takes the RGB image $\bm{I}^{RGB}$ as input and extracts multi-scale feature maps. 
For the style encoder, we first convert RGB image $\bm{I}^{RGB}$ to YUV image $\bm{I}^{YUV}$ (see Sec~\ref{sec:style_extractor}), and then apply a color mapping module to map it to a new color space that allows us to extract more discriminative style features $\bm{F}^{sty}$. 

In the UNet decoder, each decoder stage predicts an auxiliary inharmonious region mask. Moreover, we insert a style voting module after each decoder stage to guide the mask estimation in a coarse-to-fine manner and output the final mask $\hat{\bm{M}}$. By taking the $k$-th decoder stage as an example, its output inharmonious region mask $\hat{\bm{M}}^{(k)}$ and the style features  $\bm{F}^{sty}$ are sent into the style voting module to produce a voting score map $\bm{S}^{(k)}$. Additionally, we merge the multi-scale features from UNet encoder with the multi-scale features from style encoder via concatenation and $1\times 1$ convolution, leading to aggregated encoder features. Then, the voting score map $\bm{S}^{(k)}$ is concatenated with aggregated encoder features and delivered to the next decoder stage. Finally, the last decoder stage outputs the final mask $\hat{\bm{M}}$. Next, we detail our style encoder in Sec.~\ref{sec:style_extractor} and style voting module in Sec.~\ref{sec:style_voting}.


\subsection{Style Encoder}
\label{sec:style_extractor}

We adopt ResNet34 \cite{ResNet} as the backbone of style encoder to extract multi-scale features. We refer to the feature map from the last layer as style feature map, which contains task-relevant color and illumination information. We  first introduce the loss to regulate the style features, and then introduce the color mapping module which can help extract better style features.

\subsubsection{Style Feature Loss} \label{sec:style_feature_loss}
The extracted style feature map $\bm{F}^{sty}$ is expected to contain task-relevant color and illumination information, which could distinguish the inharmonious region from the background. 
Therefore, the main goal of style feature loss is to enlarge intra-region coherence and inter-region divergence.
Specifically, we use $s_{inter}$ to denote the average cosine similarity between pixel-level style features across different regions (inharmonious region and harmonious region), and $s_{intra}$ to denote the average cosine similarity between pixel-level style features within the same region (either inharmonious  or harmonious region). By denoting the ground-truth inharmonious region mask as $\bm{M}$, we define a set of inter-region pixel pairs $\mathcal{P}_{inter}=\{(p_1,p_2)| M_{p_1}\neq M_{p_2}\}$, in which $p$ is a 2D position and $M_p$ is the $p$-th entry in $\bm{M}$. Similarly, we define a set of intra-region pixel pairs  $\mathcal{P}_{intra}=\{(p_1,p_2)| M_{p_1}= M_{p_2}\}$.
By using $cos(\cdot,\cdot)$ to denote the cosine similarity between two feature vectors,
$s_{inter}$ and $s_{intra}$ are calculated as follows,

\begin{equation}
        s_{inter} = \frac{1}{|\mathcal{P}_{inter}|} \sum_{(p_1,p_2)\in \mathcal{P}_{inter}} cos(\bm{F}^{sty}_{p_1}, \bm{F}^{sty}_{p_2}), \,
    s_{intra} = \frac{1}{|\mathcal{P}_{intra}|}\sum_{(p_1, p_2)\in \mathcal{P}_{intra}} cos(\bm{F}^{sty}_{p_1}, \bm{F}^{sty}_{p_2}).
\end{equation}

We adopt the triplet loss $\ell_{sty}$ to enforce $s_{inter}$ to be smaller than $s_{intra}$ by a margin $m$: $\ell_{sty} = {\rm{max}}(s_{inter} - s_{intra} + m, 0)$, where $m$ is set to 0.5 via cross-validation. In this way, we pull close the pixel-level style features within the same region while pushing apart them across different regions. 

\subsubsection{Color Mapping Module}
To extract more discriminative style features, we insert a simple color mapping module in front of the style encoder, which converts the input image to another color space using linear color transformation. By jointly training color mapping module and style encoder supervised by the style feature loss, the intra-region coherence and inter-region divergence could be enlarged in the new color space. 

Prior to using our color mapping module, we first convert the input RGB image  $\bm{I}^{RGB} \in \mathbb{R}^{H\times W \times 3}$ to a YUV image $\bm{I}^{YUV} \in \mathbb{R}^{H\times W \times 3}$. Compared with the correlated RGB color space, YUV is a decorrelated color space, in which the luminance channel ($Y$ channel) encodes the intensity of light and the chrominance channels (U, V channels) encode the color information. Since the inharmony is caused by color/illumination discrepancy, a YUV image may exhibit the discrepancy in a better way. Moreover, we can also transform each channel independently in the decorrelated color space.

Our color mapping module applies a simple convolutional block ($3\times3$ and $7\times7$ convolutions) to the input YUV image to produce linear transformation parameters $\bm{P} = [\bm{A},  \bm{B}] \in \mathbb{R}^{H\times W\times 6}$, where $\bm{A} \in \mathbb{R}^{H\times W\times 3}$ and  $\bm{B} \in \mathbb{R}^{H\times W\times 3}$. The linear color transformation is both position-specific and channel-specific. Formally, the value $I^{YUV}_{c,p}$ in the $c$-th channel at 2D position $p$ of $\bm{I}^{YUV}$ will be mapped to a new value in the following way:
\begin{equation}
    \hat{I}_{c,p} = A_{c,p} \times I^{YUV}_{c,p} + B_{c,p},
\end{equation}
where $A_{c,p}$ and $B_{c,p}$ are the transformation parameters corresponding to the channel $c$ and 2D location $p$. After the mapping process, we can get an image $\hat{\bm{I}} \in \mathbb{R}^{H \times W \times 3}$ with each entry being $\hat{I}_{c,p}$. Then, the style encoder takes in $\hat{\bm{I}}$ and outputs a style feature map $\bm{F}^{sty} \in \mathbb{R}^{h \times w \times C}$, in which $h=\frac{H}{8}$ and  $w=\frac{W}{8}$.

\subsection{Style Voting Module}
\label{sec:style_voting}

\begin{figure}[t]
\centering
\includegraphics[scale=0.40]{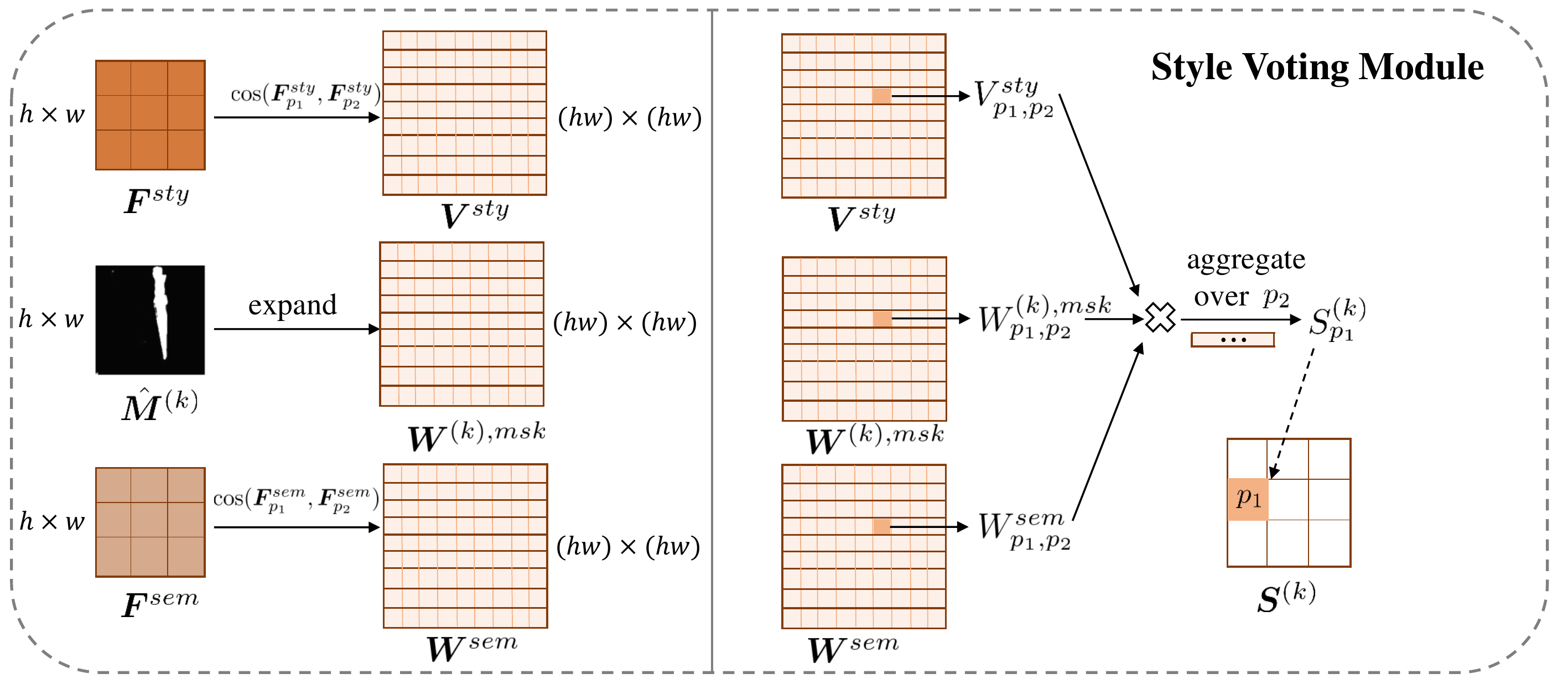}
\caption{Details of the style voting module. The left part shows constructions of style similarity matrix $\bm{V}^{sty}$, harmonious weight matrix $\hat{\bm{M}}^{(k)}$, and semantic similarity matrix $\bm{W}_{sem}$. The right part illustrates calculations of the final voting score $S^{(k)}_{p_1}$ for a specific pixel in $\bm{S}^{(k)}$.}
\label{fig:voting}
\end{figure}

Based on the extracted style feature map $\bm{F}^{sty} \in \mathbb{R}^{h \times w \times C}$, we design a novel style voting module, which produces a voting score map to indicate the harmonious pixels. We insert it after each decoder stage.  By taking the $k$-th decoder stage as an example, we use its output inharmonious region mask to select harmonious pixels as voters. The voters need to vote for similar pixels based on style features. 
After all the votings, we calculate the total score each pixel receives from all voters to be the final score for it, which indicates how likely this pixel is harmonious.  Finally, we obtain a voting score map formed by the final scores of all pixels, which highlights the harmonious regions.
The process of voting is visualized in Fig.~\ref{fig:voting}.

Our style voting module takes the style feature map $\bm{F}^{sty} \in \mathbb{R}^{h\times w\times C}$, and calculates a style similarity matrix $\bm{V}^{sty}\in \mathcal{R}^{(h\times w)\times(h\times w)}$ with the $(p_1,p_2)$-th entry $V^{sty}_{p_1,p_2}$ being the cosine similarity between the $p_1$-th pixel-level style feature and the $p_2$-th pixel-level style feature:
\begin{equation} \label{eqn:sty_sim}
    V^{sty}_{p_1,p_2} = cos(\bm{F}^{sty}_{p_1}, \bm{F}^{sty}_{p_2}),
\end{equation}
in which $V^{sty}_{p_1,p_2}$ can be viewed as the score that the $p_1$-th pixel receives from the $p_2$-th voter.

Suppose that the output inharmonious region mask from the $k$-th decoder stage is $\hat{\bm{M}}^{(k)}$, we resize $\hat{\bm{M}}^{(k)}$ to ${h\times w}$. Then, we select harmonious pixels as voters by assigning weight $(1-\hat{M}_p^{(k)})$ to the $p$-th pixel, because $\hat{M}_p^{(k)}$ is the $p$-th entry in $\hat{\bm{M}}^{(k)}$ and small $\hat{M}_p^{(k)}$ indicates reliable harmonious pixels. We replicate $\hat{\bm{M}}^{(k)}$ for $(h\times w)$ times and arrive at the weight matrix  $\bm{W}^{(k),msk}\in \mathcal{R}^{(h\times w)\times (h\times w)}$. 
Based on the weight matrix, the weighted total score that the $p_1$-th pixel receives can be represented by
\begin{eqnarray}\label{eqn:vote_score}
S_{p_1}^{(k)} = \sum_{p_2} W^{(k),msk}_{p_1, p_2} V^{sty}_{p_1,p_2}.
\end{eqnarray}
The scores of all pixels form a voting score map $\mathbf{S}^{(k)}\in \mathcal{R}^{h\times w}$, which serves as prior information to guide the UNet decoder. As shown in Fig.~\ref{fig:pipeline}, we concatenate the voting score map with the encoder features from two encoders, which are delivered to the next decoder stage.

Note that our style voting module is flexible in the weighting scheme. Besides $\bm{W}^{(k),msk}$, we can design other types of weights assigned to the voters, to select the voters which satisfy the expected property. Next, we will describe how to introduce auxiliary semantic information into the weighting scheme. 

\subsubsection{Semantic Guided Voting}
Intuitively, the objects of similar semantic categories are prone to share similar color or illumination characteristics. Hence, it would be easier to judge whether an object is harmonious or inharmonious by comparing it with other objects of similar semantic categories. For example, given an synthetic image with multiple zebras, we could tell which zebra is a composite foreground cut from another image by comparing each zebra with other zebras. 
Therefore, we explore the effect of bringing the semantic information into our voting process. When a voter is voting for a pixel, we assign a higher weight to this voter if they are semantically similar and a lower weight otherwise. 

To achieve this goal, we feed the input image to a pretrained semantic segmentation network OCRNet \cite{OCRNet} and extract the feature map from the last layer as the semantic feature map $\bm{F}^{sem}$. After resizing $\bm{F}^{sem}$ to $h\times w$, we calculate the semantic similarity matrix $\bm{W}^{sem} \in \mathcal{R}^{(h\times w) \times (h\times w)}$. Each entry $W^{sem}_{p_1,p_2}$ is the cosines similarity between each pair of pixel-level semantic features, in a similar way to (\ref{eqn:sty_sim}). We multiply $\bm{W}^{sem}$ with $\bm{W}^{(k),msk}$ as the new weight matrix, in which case the weighted total score each pixel receives can be calculated as
\begin{eqnarray}\label{eqn:vote_score_sem}
S_{p_1}^{(k)} = \sum_{p_2} W^{(k),msk}_{p_1, p_2} W^{sem}_{p_1, p_2} V^{sty}_{p_1,p_2}.
\end{eqnarray}
Then, we can replace $S_{p_1}^{(k)}$ in (\ref{eqn:vote_score}) with that in (\ref{eqn:vote_score_sem}) when constructing the voting score map. 
\subsection{Loss Function}
Following \cite{qin2019basnet,DIRL}, our mask-related loss consists of three parts: 1) the binary cross entropy loss $\ell_{bce}$; 2) the structural similarity loss $\ell_{ssim}$; 3) the intersection over union loss $\ell_{iou}$. Besides the final estimated mask $\hat{\bm{M}}$, we also supervise the auxiliary masks output from each decoder stage. Recall that we also have the style feature loss $\ell_{sty}$ to supervise the style encoder (see Section~\ref{sec:style_feature_loss}). Therefore, the total loss can be expressed as
\begin{eqnarray} \label{eqn:L_total}
    \mathcal{L} = \ell_{sty} + \ell_{bce+ssim+iou} + \sum_{k = 1}^{K}\ell_{bce+ssim+iou}^{(k)},
\end{eqnarray}
where $\ell_{bce+ssim+iou}$ is the sum of three mask-related losses. The upperscript $(k)$ indicates the output mask from the $k$-th decoder stage, and the number $K$ of decoder stages is $3$.


\section{Experiments}

\subsection{Experimental Setting}
We conduct our experiments on the image harmonization dataset iHarmony4 \cite{cong2020dovenet} following \cite{DIRL}. This dataset contains inharmonious-harmonious image pairs and the corresponding inharmonious region masks. The iHarmony4 dataset \cite{cong2020dovenet} is comprised of four sub-datasets: HAdobe5K, HCOCO, HFlickr, and HDay2Night. Following \cite{DIRL}, we only choose image pairs with area of inharmonious region smaller than 50\% of the whole image to avoid ambiguity, resulting in 64255 training images and 7237 testing images.  

Our model is implemented based on the Pytorch framework. Our optimizer is Adam with $\beta_1=0.9$, $\beta_2=0.999$, and weight decay=1e-4. We use the cosineannealing learning rate scheduler. Our model is trained for 250 epochs in total on 4 GeForce GTX TITAN X GPUs with batch size 24.  

We adopt the same evaluation metrics including Average Precision (AP), $F_1$ score, and Intersection over Union (IoU) following \cite{DIRL}. 

\subsection{Comparison with the State-of-the-art}

Besides DIRL \cite{DIRL} and MadisNet \cite{madisnet} working on inharmonious region localization, we also choose popular methods from three closely related fields for comparison. The first group is semantic segmentation networks including UNet \cite{Unet}, DeepLabv3 \cite{DeepLabv3}, HRNet-OCR \cite{OCRNet}, SegFormer \cite{xie2021segformer}. The second group is image manipulation localization methods including MantraNet \cite{MantraNet}, MAGritte \cite{MAGritte}, SPAN \cite{SPAN}. The third group is salient object detection methods including F3Net \cite{F3net}, GATENet \cite{GATENet}, MINet \cite{MINet}. For fair comparison, we also use ResNet34 for ResNet-based models.  We choose HRNet30 for HRNet-OCR and SegFormer-B3 for SegFormer for comparable model size.

\subsubsection{Quantitative Comparison}
Evaluation results of all methods are listed in Table~\ref{table:baseline}. We can see that our method achieves the best overall results and outperforms the strongest baseline MadisNet~\cite{madisnet} by a large margin. We also report the detailed results on four subdatasets, based on which our improvement mainly comes from HCOCO and Hday2night. 
We will report the comparison of computational complexity  in the Supplementary. 

\begin{table}[]
\centering
\setlength{\tabcolsep}{1mm}
\scalebox{0.7}{\begin{tabular}{@{}cccccccccccccccc@{}}

\toprule
\multirow{2}{*}{Method} & \multicolumn{3}{c}{\textbf{All}} & \multicolumn{3}{c}{HCOCO} & \multicolumn{3}{c}{HAdobe5k} & \multicolumn{3}{c}{HFlickr} & \multicolumn{3}{c}{Hday2night} \\ \cmidrule(l){2-16} 
                        & AP     & F1    & IoU    & AP     & F1     & IoU     & AP      & F1      & IoU      & AP      & F1      & IoU     & AP       & F1       & IoU      \\ \midrule
UNet &74.90 &0.6717 &64.74 &68.11 &0.5869 &56.57 &89.26 &0.8380  &80.85 &80.72 & 0.7683 &74.58 &35.74 &0.2362 &19.60 \\
DeepLabv3 &75.69 &0.6902 &66.01 & 69.09& 0.6070& 58.21 &90.20 &0.8591 &81.56 &80.01 &0.7698 &74.91 &35.87 &0.2550 &21.38 \\ 
HRNet-OCR &75.33 &0.6765 &65.49 &68.89& 0.5981 &57.69 &89.63 &0.8387 &80.98 &79.62 &0.7489 &74.55 &34.98 &0.2477 &21.34 \\ 
SegFormer &78.05 &0.7249 &66.55 &72.46 &0.6578 &58.78 & 89.43 &0.8531 &80.44 &85.19 &0.7986 &75.02 &45.16 &0.3856   &32.75 \\ \midrule
MantraNet& 64.22& 0.5691& 50.31& 56.55 &0.4811 &41.04 &81.07 &0.7510 &68.50 &67.52 &0.6302 &58.51 &28.88 &0.2019 &16.71 \\
MAGritte& 71.16& 0.6907& 60.14& 64.75& 0.6058 & 51.77& 85.50 &0.8630 & 76.36 & 75.02 & 0.7725 & 70.25 & 31.20 & 0.2549 & 17.05 \\
SPAN& 65.94& 0.5850& 54.27 & 58.41 & 0.4906 & 45.07 & 82.57 & 0.7786 & 72.49 & 69.22 & 0.6510& 62.20 & 29.58 & 0.2171 & 19.41 \\ \midrule
F3Net & 61.46& 0.5506& 47.48& 54.17& 0.4703& 40.03 & 74.31  & 0.6944 & 60.08 & 72.53 & 0.6582 & 59.31 & 30.08 & 0.2563  & 20.83 \\     
GATENet & 62.43 & 0.5296 & 46.33 & 55.07 & 0.4568 & 38.89 & 75.19 & 0.6634 & 59.18 & 74.13 & 0.6256 & 57.51 & 30.98 & 0.2174 & 19.38 \\ 
MINet & 77.51 &0.6822 & 63.04 & 71.74 & 0.6022 & 55.79 & 89.58 & 0.8379 & 77.23 & 83.86 & 0.7761 & 72.51 & 37.82 & 0.2710 & 19.38 \\ \midrule
DIRL & 80.02 & 0.7317 & 67.85 & 74.25 & 0.6701 & 60.85 & \underline{92.16} & \underline{0.8801} & \underline{84.02} & 84.21 & 0.7786 & 73.21 & 38.74 & 0.2396 & 20.11 \\ 
MadisNet(UNet) & 81.15 & 0.7372 & 67.28 & 79.02  & 0.7108  & 63.31  & 88.31 & 0.8219 & 77.41  &79.24  &0.7182 & 68.12 &49.60 & 0.3851 & 32.52 \\
MadisNet(DIRL) &85.86 &0.8022 &74.44 &83.78  &0.7741  &70.50  &\textbf{92.45}  &\textbf{0.8850} &\textbf{84.75}  &85.65  &\textbf{0.8032} &\textbf{75.49} &57.40 &0.4672 &40.47 \\ \midrule

AustNet & \underline{92.20} & \underline{0.8453} & \underline{79.63} & \underline{95.11}  & \underline{0.8866} & \underline{83.30} &89.01 & 0.8047 & 76.55 &\underline{87.72}  &0.7777 & 72.61 &\underline{74.01} &\underline{0.5554} &\underline{51.31} \\
AustNet-S & \textbf{93.01} & \textbf{0.8571} & \textbf{80.96} & \textbf{95.92}  &\textbf{0.8963} &\textbf{84.61} &89.38 &0.8113 & 76.93 & \textbf{88.21} & \underline{0.8012} &\underline{75.16} & \textbf{84.10} &\textbf{0.6438}&\textbf{60.47}

\\ \bottomrule
\end{tabular}}

\caption{Quantitative comparison with other methods on the iHarmony4 dataset including detailed results on each sub-dataset. All metrics are the larger, the better. The best method is marked in bold and the second best method is marked with underline. }
\label{table:baseline}
\end{table}

\subsubsection{Qualitative Comparison}
\begin{figure}[t]
    \centering
    \includegraphics[scale = 0.45]{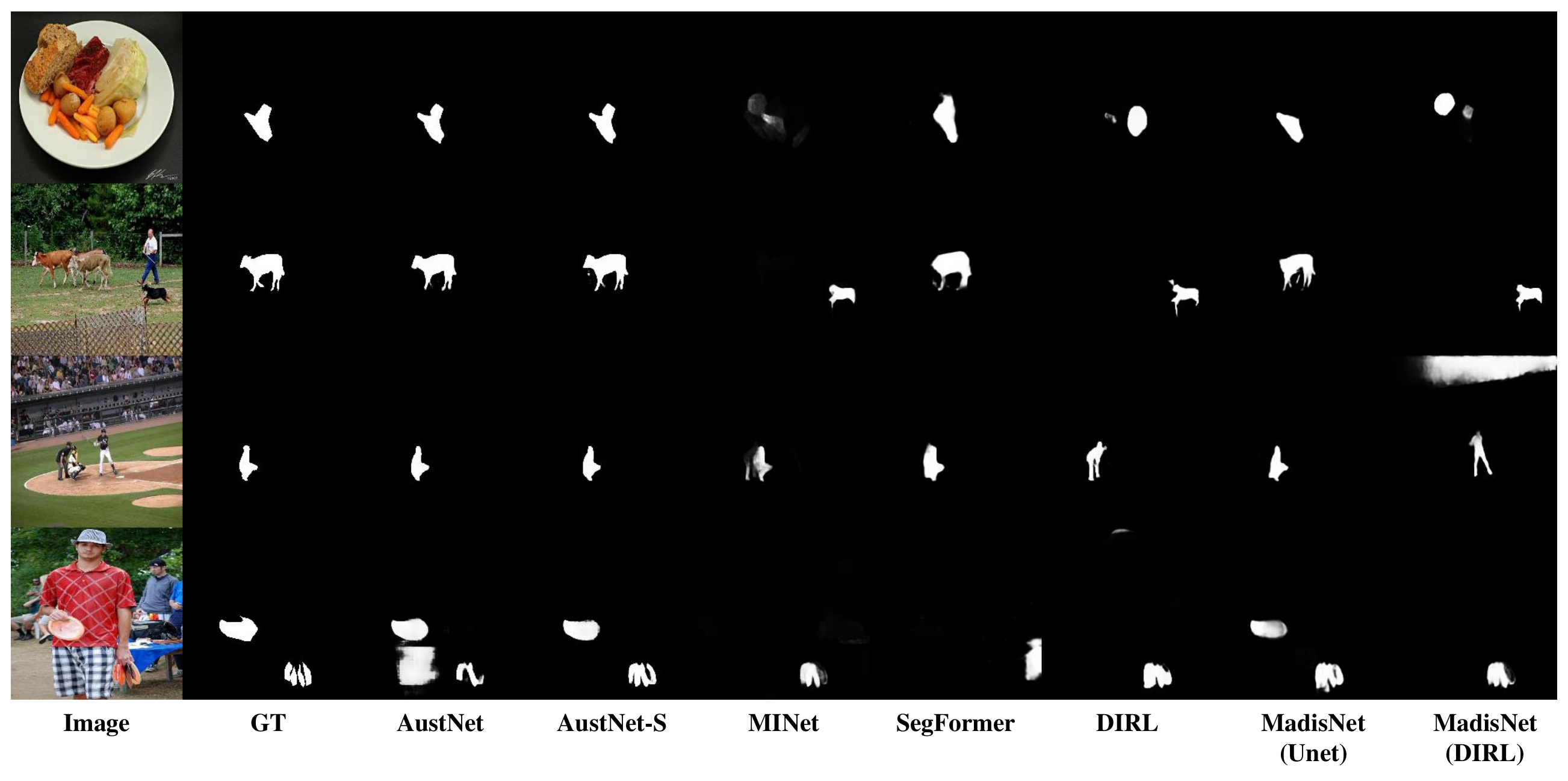}
    \caption{Qualitative comparison with baseline methods. GT is ground-truth mask.}
    \label{fig:comparisonwithsota}
\end{figure}
To visually compare our method with others, we show some visualization results of  our method and the well-behaved baselines in Fig.~\ref{fig:comparisonwithsota}. It can be seen that our method can successfully localize the inharmonious region, even in some challenging cases. More results and analyses can be found in the Supplementary.
\subsection{Ablation Studies}
In this section, we conduct comprehensive ablation studies to verify the effectiveness of our design, which are summarized in Table~\ref{table:ablation}. The first row only contains a simple encoder-decoder branch with RGB image as input. In row 2, we change the input image to YUV color space and observe performance improvement, which shows that suitable color space is beneficial for the inharmonious region localization task. For row 3, we simply add a style encoder without color mapping module or style feature loss, and concatenate the multi-scale encoder features from two encoders as in our method.  This leads to minor improvement compared with row 2, implying that the style feature should be utilized in a better way. 

\begin{table}
\centering
\scalebox{0.8}{
\begin{tabular}{@{}ccccccccc@{}}
\toprule
\# & \multicolumn{5}{c}{Components}                                                                                   & \multicolumn{3}{c}{Evaluaion Metrics}                              \\ \midrule
   & Input                & Color-mapping        & Voting               & $\ell_{sty}$                    & Semantic             & AP                   & F1                   & IoU                  \\ \midrule
1 & RGB &  &  &  &  & 73.45  & 0.6330 &56.76    \\
2 & YUV& & & & &76.28& 0.6511 & 59.24    \\
3 & RGB+YUV &  &  & & & 76.45 & 0.6573& 59.73     \\
4 & RGB+YUV & & only aux mask & & & 77.32 & 0.6569& 59.81    \\
5 & RGB+RGB & & \checkmark & \checkmark & &75.57 & 0.6685&  60.24   \\
6 & RGB+YUV& & \checkmark &\checkmark& &79.10 &0.6986& 64.37    \\
7 & RGB+YUV & & \checkmark& \checkmark & \checkmark &79.56 & 0.7242 & 66.42   \\
8 & RGB+YUV & \checkmark & &  & & 86.17 & 0.7741 & 70.61    \\
9 & RGB+YUV& \checkmark & & \checkmark & & 86.97 & 0.7826& 71.98    \\
10 & RGB+RGB& \checkmark & & \checkmark& & 76.78 & 0.6599 & 59.65    \\
11 & RGB+YUV& \checkmark & \checkmark& \checkmark & &92.01 & 0.8477& 78.78    \\
12 & RGB+YUV& \checkmark & \checkmark & \checkmark& \checkmark          
& \textbf{93.01}& \textbf{0.8600} &\textbf{81.14}     \\ \bottomrule
\end{tabular}}
\caption{Ablation study of key components in our method.}
\label{table:ablation}
\end{table}

Based on row 3, we add our style voting module in row 6, the performance gain proves the effectiveness of our style voting module. Adding semantic information in row 7 further boosts the result. We change the YUV color space in row 6 to RGB, and the obtained results in row 5 indicate the advantage of YUV color space.
To ensure that the major improvement is not brought by predicting auxiliary masks in each decoder stage, we experiment to only predict auxiliary masks without style voting module in row 4. 

In row 8-12, we conduct experiments with color mapping module. From row 8, we see that our color mapping module significantly advances the performance and adding style feature loss in row 9 brings further improvement. In row 10, we replace the input YUV image with RGB image and the performance drops sharply. The results in row 11 and 12 demonstrate the effectiveness of (semantic-guided) style voting map. 

\subsection{Experiments on Multiple Inharmonious Regions}\label{sec:multi foreground}
Images in the iHarmony4 \cite{cong2020dovenet} dataset mainly contain a single inharmonious region, but in real-life scenario, it is possible that there are several separate  inharmonious regions in one image and each inharmonious region may also be different in terms of color and illumination. To investigate the ability of our model to detect multiple inharmonious regions, we build a set of test images with multiple disjoint inharmonious regions based on the HCOCO subset of iHarmony4.
Specifically, real images in HCOCO may have different inharmonious image pairs with different manipulated foregrounds. Thus, we combine these inharmonious images corresponding to a single real image to construct a test set with multiple inharmonious regions. This test set contains 19482 images in total, with the number of inharmonious regions ranging from 2 to 9. We compare our AustNet and AustNet-S with the strongest baseline MadisNet \cite{madisnet} on this test set. The detailed quantitative results and visualization results are left to the Supplementary. 

\section{Conclusions}

In this work, we focus on the essence of inharmonious region localization task and extract discriminative style features. Centering around the style features, we have proposed a novel color mapping module and a novel style voting model to help localize the inharmonious region. We have also verified the effectiveness of utilizing semantic information in the voting process. Our method significantly outperforms the existing methods.

\section*{Acknowledgement}

The work was supported by the Shanghai Municipal Science and Technology Major/Key Project, China (2021SHZDZX0102, 20511100300) and  National Natural Science Foundation of China (Grant No. 61902247).
\bibliography{final}

\end{document}


\maketitle

In this supplementary, we compare our model with other baselines on model size and speed in Sec.~\ref{sec:model_size}. We conduct more experiments to explore and discuss the effectiveness of our design in Sec.~\ref{sec:additional experiments}. Last, we test our model on synthetic images with multiple inharmonious regions in Sec.~\ref{sec:multi foreground}.


\section{Comparison of Model Size and Speed} \label{sec:model_size}

In this section, we compare the number of parameters, GFlops, and inference time between our methods and 4 competitive baselines in Table~\ref{table:parametercomparison}. All the evaluations are conducted on a single GTX TITAN X GPU. We can see that the number of parameters and the inference time of our AustNet are comparable with others, though the GFlops is larger. It can also be seen that the choice of semantic segmentation model of our AustNet-S largely determines the number of parameters and inference time. Therefore, we can choose between AustNet and AustNet-S according to our need when dealing with performance-speed/parameter trade-off.
\begin{table}[t]
\centering
\begin{tabular}{|c|c|c|c|}
\hline
          & Number of Parameters & GFlops & Inference Time \\ \hline
SegFormer & 48.28M               & 30.97  & 61.36ms        \\ \hline
MINet     & 68.28M               & 116.01 & 63.95ms        \\ \hline
DIRL      & 53.46M               & 104.67 & 63.80ms        \\ \hline
MadisNet(UNet)      &  59.94M              & 82.90 & 45.60ms      \\ \hline
MadisNet(DIRL)      &  56.69M    & 105.12  & 71.80ms \\ \hline
AustNet   & 50.05M               & 167.01 &  70.90ms              \\ \hline
AustNet-S & 120.61M              & 198.70  & 148.30ms         \\ \hline
\end{tabular}
\caption{Comparison of number of parameters, GFlops, and inference time between our model and other baselines.}
\label{table:parametercomparison}
\end{table}

\section{Additional Experiments} \label{sec:additional experiments}

\subsection{More Qualitative Comparison }
\begin{figure}
    \centering
    \includegraphics[scale=0.77]{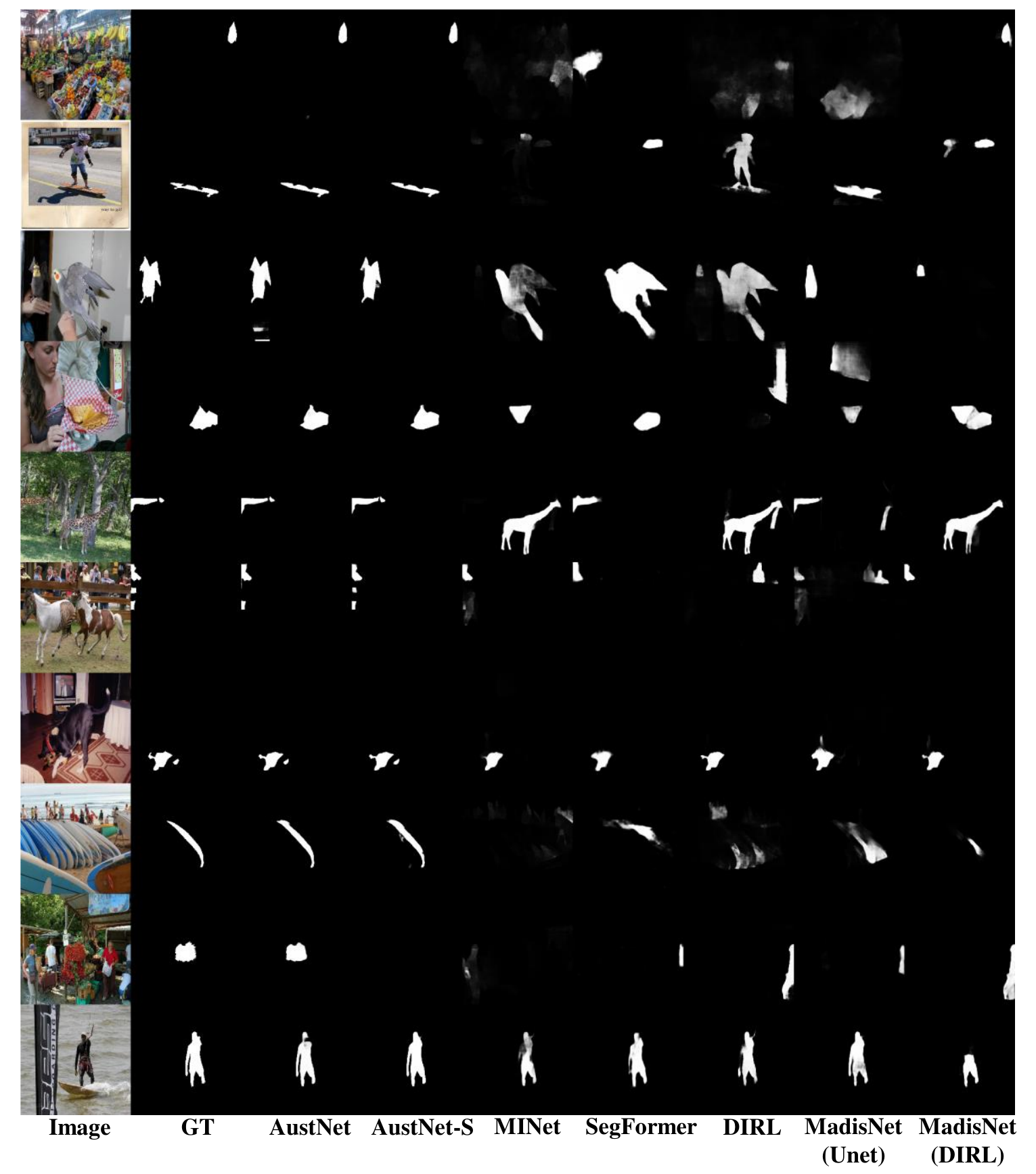}
    \caption{Qualitative comparison with baseline methods. GT is ground-truth mask.}
    \label{fig:comparisonwithsota_more}
\end{figure}

We provide more qualitative comparison between our method with other baselines in Fig~\ref{fig:comparisonwithsota_more}. Our method can accurately detect the inharmonious regions in challenging cases.

\subsection{Study on Style Feature}
\begin{table}[t]
\centering
\begin{tabular}{|c|c|c|c|}
\hline
& RGB+YUV & RGB+YUV+Color-mapping & RGB+YUV+Color-mapping + $\ell_{sty}$ \\ \hline
$s_{inter}$ & 0.3503        &    0.3435                   &   0.1588                         \\ \hline
$s_{intra}$ & 0.4792        &         0.6212              &   0.7402                         \\ \hline
\end{tabular}
\caption{Study on the discriminativeness of extracted style feature in terms of $s_{inter}$ and $s_{intra}$.}
\label{table:colormappingstudy}
\end{table}

To investigate the discriminativeness of extracted style feature, we calculate the average inter-region feature similarity $s_{inter}$ and average intra-region feature similarity $s_{intra}$ over test images for models with and without our color-mapping module and $\ell_{sty}$. The results are shown in Table~\ref{table:colormappingstudy}. We observe that when our color mapping module or $\ell_{sty}$ is missing, $s_{inter}$ and $s_{intra}$ are close, indicating that the style features become less discriminative. The discrepancy between $s_{inter}$ and $s_{intra}$ is enlarged after adding the color mapping, and further enlarged after adding $\ell_{sty}$, which means that our extracted style features are discriminative and informative enough to separate the inharmonious region from the background.

\subsection{Study of the Voting Process in Different Stages}
\begin{figure}[t]
\centering
\includegraphics[scale=0.35]{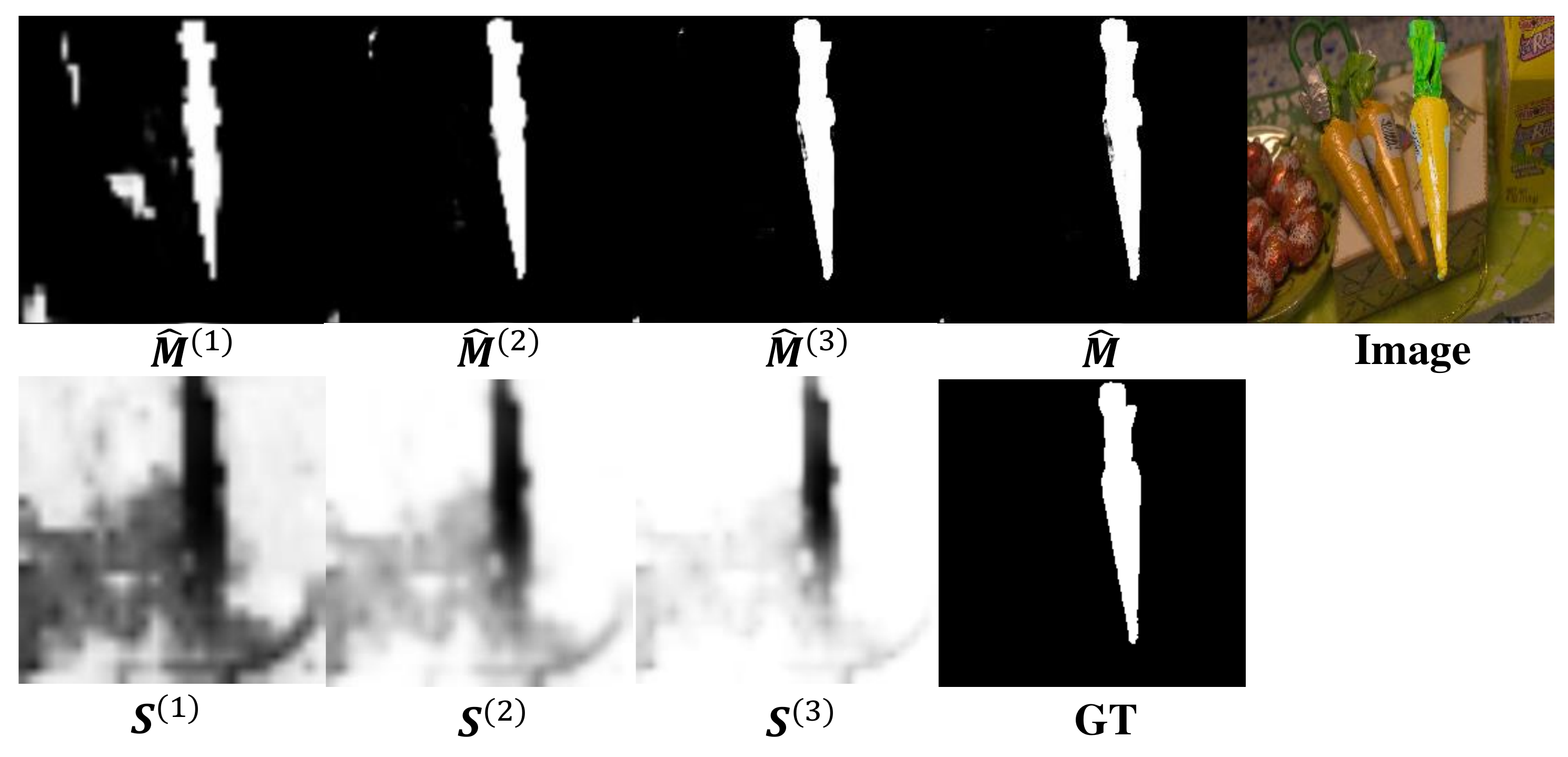}
\caption{Visualization of the voting score map $\bm{S}^{(k)}$ and auxiliary inharmonious region mask $\hat{\bm{M}}^{(k)}$ in each decoder stage from our AustNet. Brighter region in $\bm{S}^{(k)}$ means higher score.}
\label{fig:process}
\end{figure}

We visualize the voting score map and estimated inharmonious region mask in each decoder stage of our AustNet in Fig.~\ref{fig:process}. We can see that initial inharmonious region mask contains some mis-detected regions, and the voting score map is also not very accurate with some harmonious pixels receiving relatively low scores. However, as the decoding process goes, we introduce the estimated inharmonious region mask into the voting process, making both voting score maps and inharmonious region masks more and more accurate.

\subsection{Benefit of Semantic Information in the Voting Process}
\begin{figure}
\centering
\includegraphics[scale=0.53]{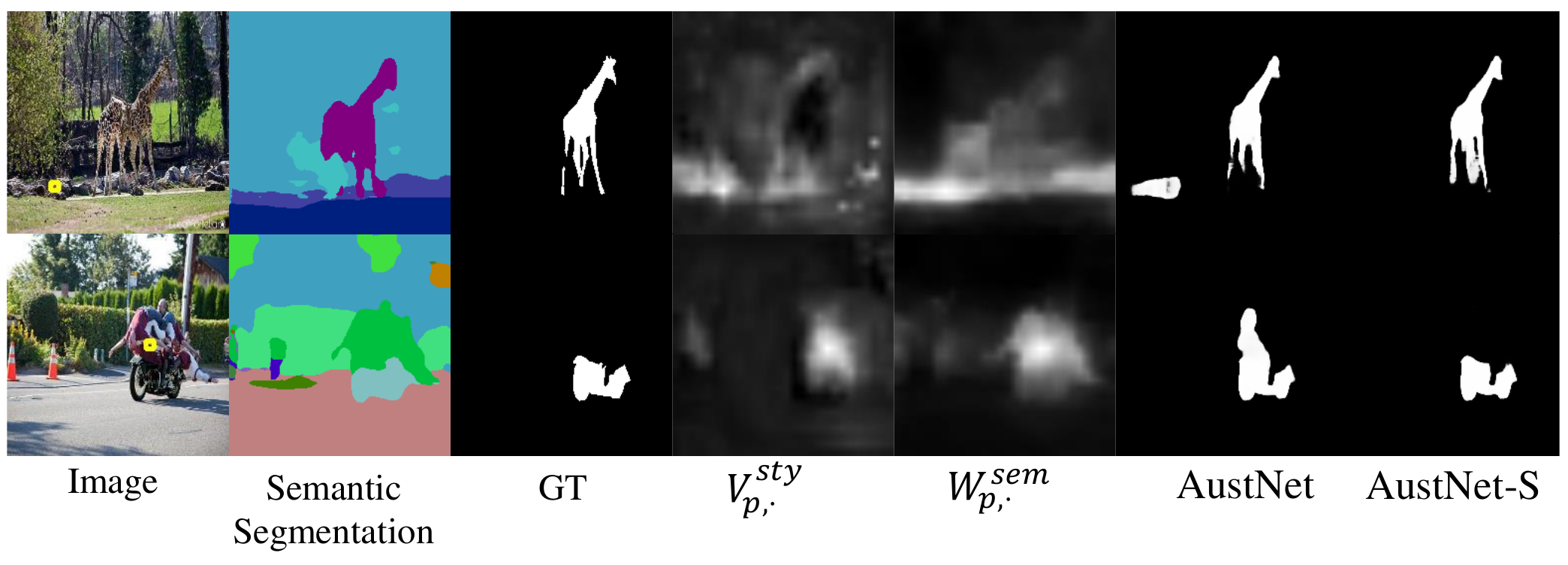}
\caption{From left to right: input image, predicted segmentation mask, ground-truth inharmonious region mask, style similarity map $\bm{V}^{sty}_{p,\cdot}$ and semantic similarity map $\bm{W}^{sem}_{p,\cdot}$ of the specified point (yellow box in the input image), predicted inharmonious region mask from our AustNet and AustNet-S.}
\label{fig:semantic_weight}
\end{figure}

When some harmonious objects have very different color style from the main background in the image,  AustNet may mis-detect them as inharmonious region since the style features may not be perfectly learned. In this case, our AustNet-S with semantic information is designed to remedy this problem. 
To verify the benefit of semantic information in the voting process, we show several examples of the comparison between AustNet and AustNet-S in Fig.~\ref{fig:semantic_weight}. For the first case, both our AustNet and AustNet-S successfully localize the inharmonious giraffe, but AustNet mis-detects a part of cobblestone as inharmonious. We choose a mis-detected point $p$ (marked with yellow box), and visualize the style similarity matrix and semantic similarity matrix at point $p$. As we can see from the style similarity map $V_{p,\cdot}^{sty}$, since the main background of grassland and woods is green, only a small part of cobblestone region has similar style features to this point $p$. Therefore, in the style voting module of AustNet, this point receives low scores from many points in the green background, which causes it to be viewed as inharmonious. However, with semantic information in the style voting module, the weights of score it receives from other cobblestone region will be large (as shown in $W_{p,\cdot}^{sem}$), so it would receive a relatively high score and not be classified as inharmonious. Similarly, in the second example, a person is also mis-detected as inharmonious by AustNet. After adding semantic information, the mis-detected point would receive high score from other people in the image, making the final estimation of AustNet-S only contain the motorcycle part.

\section{Experiments on Multiple Inharmonious Regions}\label{sec:multi foreground}
\begin{figure} [t]
    \centering
    \includegraphics[scale=0.35]{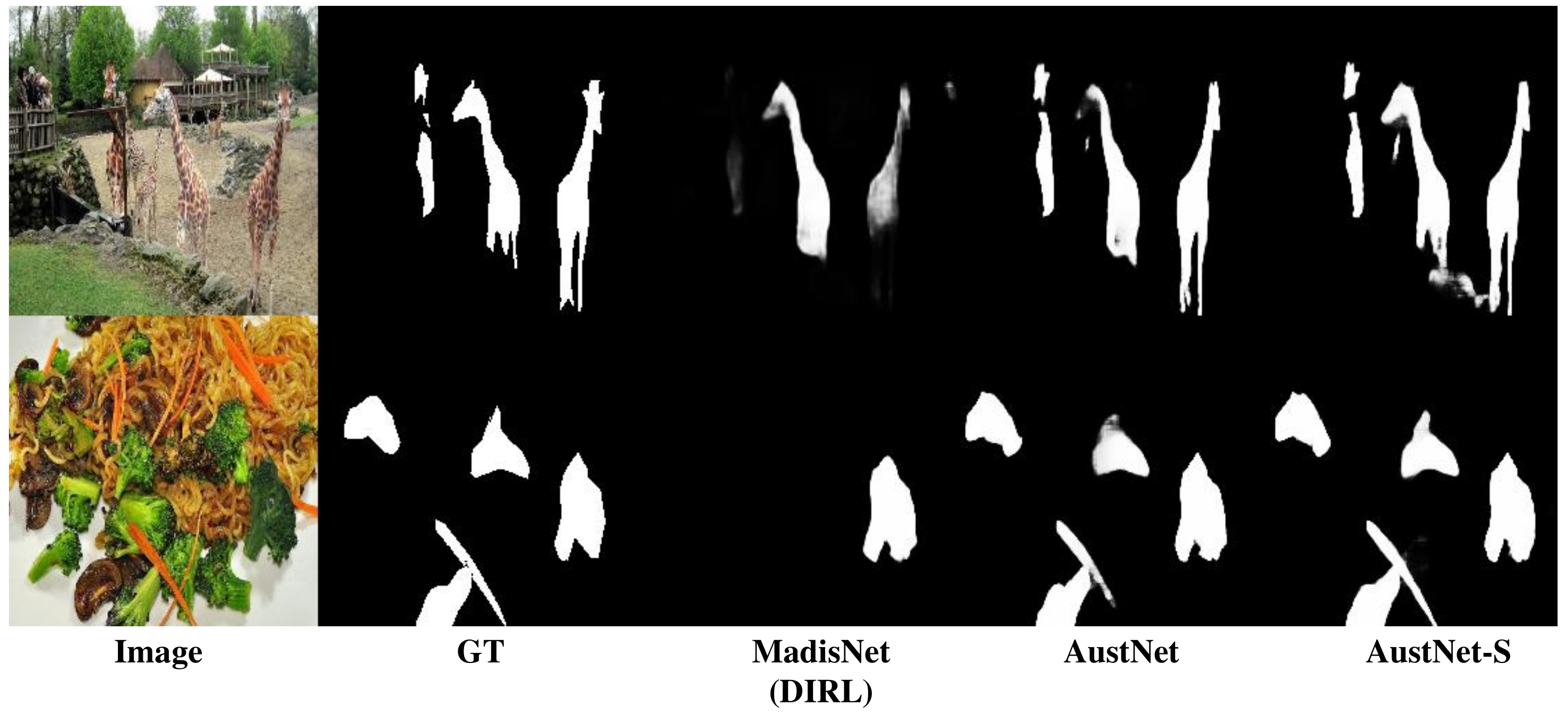}
    \caption{Visualization of the results on test images with multiple disjoint inharmonious regions.}
    \label{fig:multi_foreground}
\end{figure}

As introduced in the main paper,  we build a set of test images with multiple disjoint inharmonious regions based on the HCOCO subset of iHarmony4. This test set contains 19482 images in total, with the number of inharmonious regions ranging from 2 to 9.
We compare our AustNet and AustNet-S  with the baseline MadisNet(DIRL) \cite{madisnet}. The evaluation results (AP, $F_1$, IoU) are (77.39, 0.6761, 54.03) for MadisNet(DIRL) and (87.86, 0.7828, 66.63) for our AustNet and (89.22, 0.7972, 68.71) for AustNet-S. Some visualization results are shown in Fig.~\ref{fig:multi_foreground}.

\bibliography{supp}